\documentclass[preprint,12pt,authoryear]{elsarticle}

\usepackage{lineno}
\usepackage{times}
\usepackage{url}
\usepackage{latexsym}
\usepackage{xcolor} 
\usepackage{soul}
\usepackage{microtype}
\usepackage{graphicx}
\usepackage{booktabs}
\usepackage{microtype}
\usepackage{latexsym}
\usepackage{arydshln}
\usepackage{textcomp}
\usepackage{float}
\usepackage{multirow}
\usepackage{enumitem}
\usepackage{booktabs,fixltx2e}
\usepackage{amssymb,mathtools}
\usepackage{amsmath} 
\usepackage{eqnarray,amsmath} 
\usepackage[unicode=false]{hyperref}
\usepackage{microtype}
\usepackage{subfig}
\usepackage{pbox}
\usepackage{pifont}
\usepackage{tikz}
\usepackage{tikz-cd}
\usetikzlibrary{arrows.meta, positioning}

\definecolor{LightGoldenrod11}{rgb}{0.98, 0.98, 0.82}
\usepackage{amsthm}

\theoremstyle{definition}

\theoremstyle{remark}

\begin{document}

\nocite{*} 

\title{VisioPhysioENet: Visual Physiological Engagement Detection Network}

\author[1]{Alakhsimar Singh\corref{equal}}
\author[1]{Kanav Goyal\corref{equal}}
\author[1]{Nischay Verma\corref{equal}}
\author[2]{Puneet Kumar}
\author[3,2]{Xiaobai Li}
\author[1]{Amritpal Singh\corref{cor1}}
\ead{apsingh@nitj.ac.in}



\cortext[cor1]{Corresponding author}
\cortext[equal]{Contributed equally.}

\address[1]{Department of Computer Science and Engineering, NIT Jalandhar, India}
\address[2]{Center for Machine Vision and Signal Analysis, University of Oulu, Finland}
\address[3]{State Key Lab of Blockchain and Data Security, Zhejiang University, China}

\begin{abstract}
    This paper presents VisioPhysioENet, a lightweight multimodal system for detecting learner engagement by jointly leveraging visual and physiological signals. Unlike most existing works that rely primarily on visual cues, VisioPhysioENet integrates physiological signals non-invasively from standard webcams, without the need for wearable sensors and ensuring suitability for real-life, scalable deployment. The framework employs a two-level feature extraction process: for visual features, Dlib detects facial landmarks and OpenCV computes metrics such as head pose and gaze direction; for physiological features, the face recognition library (built on Dlib) isolates the forehead region for remote photoplethysmography (rPPG) signals extraction using the plane-orthogonal-to-skin method. Extracted features are fused via a late fusion strategy combining AdaBoost (visual) and Random Forest (physiological) classifiers, with stacked logistic regression for final prediction. Evaluation on the DAiSEE dataset shows VisioPhysioENet achieves an accuracy of 63.09\%, outperforming the only other visual+physiological method by 8.6\%, while maintaining low computational overhead. This work demonstrates the potential of convenient multimodal engagement detection, offering a pioneering yet extensible pathway for improving robustness, accuracy, and real-world applicability. Code is available at \href{https://github.com/MIntelligence-Group/VisioPhysioENet}{github.com/MIntelligence-Group/VisioPhysioENet}.
\end{abstract}


\maketitle
\section{Introduction}\label{sec:intro}
Engagement detection in educational settings is essential for optimizing learning outcomes \cite{alkabbany2019measuring}. Traditional methods, often limited by their reliance on a single data type, may not fully capture the complexities of learner engagement \cite{kumar2024measuring, chen2019faceengage}. These single-modality approaches typically utilize visual information, physiological data, or subjective self-reports independently, thus inherently missing the nuanced interplay between cognitive, emotional, and behavioral dimensions of engagement. Additionally, they frequently encounter practical limitations such as susceptibility to environmental variations like illumination changes, occlusions in video-based analysis, or artifacts arising from physical movement in wearable physiological sensors. Such factors significantly restrict their robustness and applicability in diverse and dynamic learning scenarios, particularly in real-world classrooms or remote learning environments.

Recent studies advocate multimodal engagement detection, integrating various complementary data types to overcome these inherent limitations of unimodal approaches \cite{kumar2022hybrid, sumer2021multimodal,kumar2025multimodal}. Although multimodal systems combining visual and physiological signals have shown improved accuracy and comprehensiveness \cite{abedi2021improving, selim2022students}, several persistent challenges remain unaddressed. These include class imbalance, wherein particular engagement levels such as ``highly engaged'' or ``highly disengaged'' occur less frequently, complicating robust classifier training \cite{xie2023student, mehta2022three}. Many existing multimodal systems also exhibit limited generalizability across datasets and learning contexts, often relying on highly specialized deep-learning pipelines that demand extensive computational resources and large volumes of labeled data. Furthermore, most prior work has heavily favored visual modalities, with relatively little emphasis on fully exploiting physiological signals, particularly remote cardiovascular measures, despite their proven relevance in cognitive and affective state assessment \cite{kumar2024measuring, sarkar2020self}.

To address these gaps, we propose VisioPhysioENet, a lightweight, non-invasive, and real-life–suitable multimodal framework that jointly exploits visual cues and remote physiological (rPPG) signals from standard webcams. By removing the need for intrusive wearable sensors, the method is designed for convenient deployment in authentic learning environments, including large-scale or hybrid classrooms. The architecture is computationally efficient, avoiding overly complex deep models while still capturing complementary modality-specific strengths.

For visual feature extraction, the framework employs Dlib's 68-point facial landmark detector and OpenCV functions to compute key metrics such as Eye Aspect Ratio (EAR), head pose angles (pitch, yaw, roll), and gaze direction \cite{dlib2016davis, bradski1999opencv, gao2025gaze}. These features are transformed into interpretable categorical indicators (eye openness, gaze stability, head orientation) closely linked to cognitive and emotional states. In parallel, rPPG signals are extracted from the forehead region using the Plane-Orthogonal-to-Skin (POS) method \cite{ghosh2015remote} with the face\_recognition library \cite{face_recognition}, yielding physiological descriptors such as heart rate, peak-to-peak intervals, and cumulative systolic/diastolic peaks. Compact, single-value features are derived to reduce computational demands while maintaining discriminative power.

The two modalities are processed separately with machine learning models selected for their suitability to each feature type: AdaBoost for visual cues, which excels at refining weak learners on structured categorical data, and Random Forest for physiological descriptors, known for robustness to multivariate signals. Late fusion via stacked logistic regression combines their outputs, enhancing predictive accuracy and interpretability.

Extensive evaluation on the DAiSEE dataset \cite{gupta2016daisee} comprising 9,068 annotated videos demonstrates that VisioPhysioENet achieves an accuracy of 63.09\%, surpassing all unimodal baselines and outperforming by 8.6\% the only other reported model that integrates both visual and physiological modalities. The system’s low computational overhead, combined with its remote sensing capability, makes it well-suited for continuous monitoring in real-world educational scenarios. Moreover, as a pioneering yet extensible design, VisioPhysioENet offers a practical pathway for future multimodal engagement detection research, where additional modalities or adaptive learning strategies can be incorporated without sacrificing its lightweight, real-life–ready nature. The major contributions of this paper are listed as follows:\vspace{.03in} 
\begin{itemize}[]
    \item Proposal of VisioPhysioENet, a lightweight and computationally efficient multimodal engagement detection system that integrates visual cues with rPPG signals from standard webcams without the need for wearable or intrusive devices.
    \item Design of a robust two-level feature extraction pipeline, using Dlib and OpenCV for facial metrics, and the POS method for rPPG signal analysis.
    \item Comprehensive evaluation of early and late multimodal fusion strategies, with the proposed late fusion (AdaBoost + Random Forest + stacked logistic regression) achieving superior performance over unimodal and existing V+P methods.
    \item Experimental validation on the DAiSEE dataset, outperforming the only other visual+physiological method by 8.6\%, supported by ablation studies that address choice of suitable features, machine learning model, fusion method and computational efficiency for scalable deployment.
\end{itemize}

\section{Literature Review}\label{sec:related_works}
The detection and analysis of learner engagement have advanced significantly, progressing from simple single-modal methods to sophisticated multimodal systems. These advancements aim to address the inherent complexities and dynamic nature of engagement detection in real-world educational scenarios, including both formal classroom environments and informal learning settings.

\subsection{Unimodal Engagement Detection}
Initial research in engagement detection predominantly focused on single data modalities, particularly visual or physiological signals. In the visual domain, Abedi et al. \cite{abedi2021improving} utilized ResNet with Temporal Convolutional Networks (TCNs) to extract spatiotemporal features from video datasets such as DAiSEE. Their approach emphasized the value of capturing temporal dynamics in facial expressions and gaze patterns for engagement prediction. Similarly, Selim et al. \cite{selim2022students} leveraged EfficientNet combined with Long Short-Term Memory (LSTM) networks for video-based engagement detection, effectively demonstrating the strength of sequential modeling techniques for capturing temporal dependencies. However, these visual-based approaches often struggled with limitations such as imbalanced datasets, which adversely affected classifier training and generalizability.

Physiological signals have also been independently explored to assess engagement due to their strong correlations with cognitive and emotional states. Sarkar et al. \cite{sarkar2020self} showcased the potential of using electrocardiogram (ECG) data for emotion classification, directly linked to learner engagement. Gupta et al. \cite{gupta2023multimodal} expanded upon this by integrating facial cues with physiological indicators like heart rate variability, achieving a more comprehensive assessment of learners' internal states. These contributions underscore the critical role of physiological data in capturing subtle, non-visible aspects of engagement, such as cognitive load and emotional arousal, that purely visual methods might overlook.

\subsection{Multimodal Engagement Detection}
To overcome the limitations of unimodal methods, researchers have increasingly explored multimodal fusion strategies, particularly those integrating visual and physiological modalities. Early and late fusion have emerged as primary approaches for combining diverse data streams effectively. Xie et al. \cite{xie2023student} and Mehta et al. \cite{mehta2022three} investigated both approaches, with early fusion combining raw or minimally processed features from different modalities into a unified representation to leverage complementary information early in the feature space, while late fusion integrates outputs or predictions from modality-specific classifiers to provide robust decision-making by combining independently derived inferences. More sophisticated integration strategies, such as stacked generalization and ensemble learning, have also been employed to systematically combine multiple classifier outputs, handling uncertainty and variability across modalities effectively \cite{grinsztajn2022tree}. These methods have consistently improved prediction accuracy, robustness, and generalizability, particularly in complex educational environments with data variability and environmental challenges.

Within this context, the synthesis of visual and physiological data streams has produced advanced multimodal systems capable of nuanced behavior and engagement analysis. Yang et al. \cite{yang2023multimediate} and Liao et al. \cite{liao2021deep} demonstrated the advantages of modality fusion in capturing complementary engagement indicators, achieving increased robustness and predictive accuracy. Savchenko et al. \cite{savchenko2022classifying} introduced streamlined Convolutional Neural Network (CNN) architectures optimized for multimodal engagement tasks, reducing computational complexity while maintaining high accuracy, while Abedi et al. \cite{abedi2024engagement} validated these strategies for real-time applications, offering practical insights into their operational performance and feasibility.

\subsection{Advances in Model Architectures and Generalizability}
Recent innovations in deep learning architectures have further advanced multimodal engagement detection. Vedernikov et al. \cite{vedernikov2024tcct} and Dubbaka et al. \cite{dubbaka2020detecting} refined deep neural networks specifically tailored to educational settings, improving model adaptability and generalization to diverse learning environments. Additionally, Copur et al. \cite{copur2022engagement} and Singh et al. \cite{singh2023have} extended engagement analysis into informal and digital learning environments, showcasing the flexibility and adaptability of multimodal models beyond traditional classroom settings. These studies highlight a growing emphasis on deploying engagement detection systems that perform reliably under varying conditions, thus broadening their applicability across educational contexts.

Machine learning models have also continued to play a pivotal role, with ensemble methods and traditional classifiers frequently employed to address persistent challenges such as data imbalance, computational efficiency, and robustness. Sherly et al. \cite{sherly2024fostering}, Hasnine et al. \cite{hasnine2021students}, and Sharma et al. \cite{sharma2022student} implemented various machine learning algorithms, such as random forests and boosting techniques, demonstrating that combining multiple weak learners significantly enhances predictive accuracy. These ensemble-based approaches proved particularly effective in overcoming class imbalance issues, further underscoring their relevance in engagement detection research.

\subsection{Real-Time Engagement Detection}
Real-time detection of engagement represents another critical research direction. Chen et al. \cite{chen2019faceengage} and Lee et al. \cite{lee2022predicting} specifically addressed the challenges inherent to deploying systems capable of continuous monitoring and rapid response in real-world educational scenarios. They emphasized the importance of developing lightweight, computationally efficient models that can perform reliably and swiftly under diverse environmental and technological conditions. Similarly, Gupta et al. \cite{gupta2023facial} proposed advanced deep learning-based systems tailored explicitly for dynamic environments, highlighting the feasibility and necessity of real-time monitoring and adaptive response capabilities within multimodal engagement detection frameworks. \vspace{.1in}

Building upon these collective advancements, this work addresses the aforementioned gaps in the existing literature. While prior multimodal frameworks have demonstrated the potential of combining visual and physiological modalities, most either rely on computationally heavy deep-learning pipelines or on contact-based physiological sensors that are intrusive and impractical for large-scale deployment. In contrast, we propose VisioPhysioENet, a lightweight, non-invasive multimodal engagement detection framework that robustly fuses visual cues with rPPG signals captured from standard webcams. By converting unstructured video into structured numeric features, our approach enables effective dimensionality reduction and leverages traditional machine learning models for fast, resource-efficient inference \cite{grinsztajn2022tree}. Through carefully designed evaluations of both early and late fusion strategies, we show that our late fusion approach not only achieves superior interpretability but also outperforms the only other reported visual+physiological method by 8.6\% on the DAiSEE dataset. This improvement, combined with the framework’s scalability, low computational overhead, and suitability for real-time deployment, directly addresses long-standing challenges of computational complexity, class imbalance, and environmental variability, offering a practical and extensible solution for real-world educational contexts.

\section{VisioPhysioENet}\label{sec:method}
\begin{figure}[!h]
\centering
\includegraphics[width=1\textwidth]{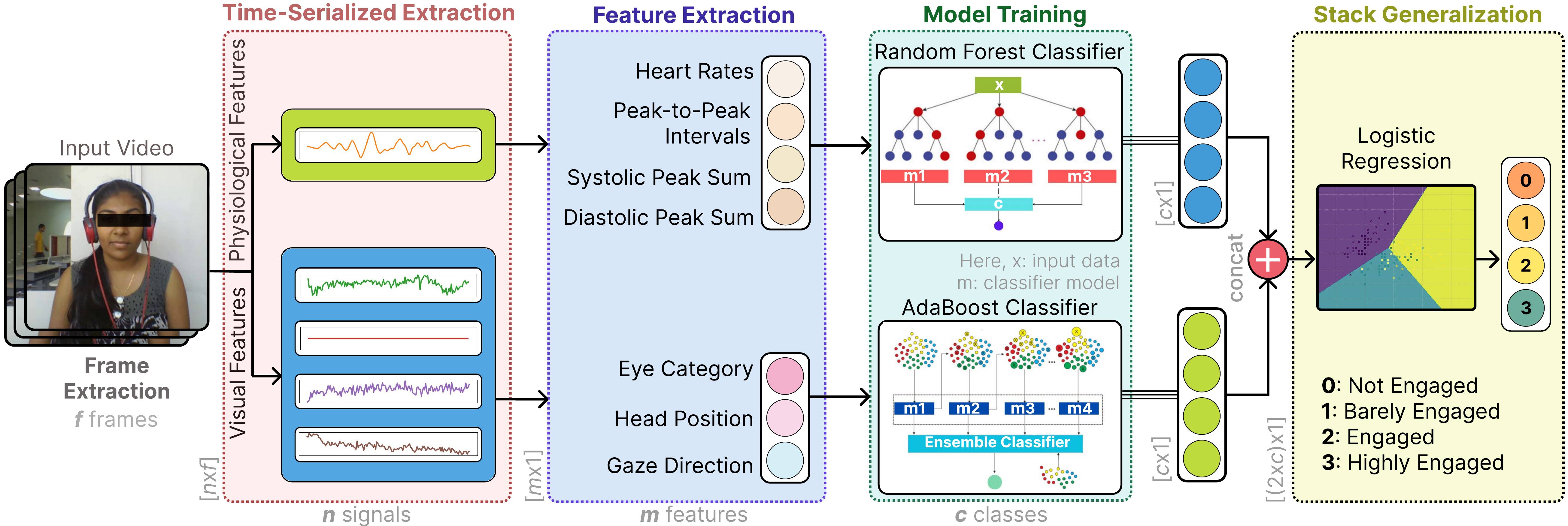} 
\caption{Schematic architecture of VisioPhysioENet encompassing feature extraction modules for visual and physiological modalities along with the fusion module.}\vspace{-.2in}
\label{fig:archi}
\end{figure}

VisioPhysioENet is a lightweight, multimodal engagement detection framework designed to operate in real-world settings by efficiently combining visual and physiological features through a structured feature extraction and fusion pipeline. Its architecture is illustrated in Fig.~\ref{fig:archi} and described in detail in the following sections.

\subsection{Visual Feature Extraction Phase}\label{vis}
The visual feature extraction model works in two levels, denoted as $\pounds_i^V$ for $i \in (1,2)$, to extract features from video data ($V$) by processing frames ($f_i$) sequentially.
In the first level ($\pounds_1^V$), the model extracts raw features such as facial landmarks ($p_1$ to $p_{68}$), Eye Aspect Ratio ($\xi$), and head orientation parameters like pitch ($\theta_x$) and yaw ($\theta_y$). The second level ($\pounds_2^V$) derives higher-level categorical features from the outputs of $\pounds_1^V$. These features are determined based on threshold values and include eye gaze direction ($\varphi$), head position category ($\hbar$), and eye openness category ($\epsilon$). The next sections describe each stage in more detail.


\subsubsection{Level $1$ ($\pounds_1^V$) Visual Feature Extraction }
This process involves extracting facial landmarks using a pre-trained 68-point Facial Landmark (FLM) detection model from the Dlib library. The facial features considered in this experiment include eye openness, EAR ($\xi$), and head orientation parameters such as pitch ($\theta_x$), yaw ($\theta_y$), and roll ($\theta_z$). These features are extracted based on the detected landmarks.

\begin{eqnarray}\scriptsize \label{eq1}
 \forall i \Big\{(f_i\ in\ V) \big\{\pounds_1(f_i)\big\}\Big\}\xrightarrow[Dlib]{[FLM]}\bigg\{\Big\{p_1-p_{68} \Big\}\bigg\};\bigg\{\Big\{ \theta_x, \theta_y, \theta_z \Big\}; \Big\{\xi\Big\} \bigg\}
 \end{eqnarray}

 
 The EAR ($\xi$) is calculated using the facial landmark points of both the left ($\xi_L$) and right eyes ($\xi_R$), and then their mean is taken.
\begin{eqnarray}
\xi\xleftarrow[/2]{\sum}\begin{cases}
\xi_{L}=\frac{\| p_{37} - p_{41} \|+ \| p_{38} - p_{40} \|}{2\| p_{36} - p_{39} \|} & \\
& \\
 \xi_{R}=\frac{\| p_{43} - p_{47} \|+ \| p_{44} - p_{46} \|}{2\| p_{42} - p_{45} \|}& 
\end{cases} 
\end{eqnarray}
where $\xleftarrow[/2]{\sum}$ denotes the computation of the mean and assignment operator. 

For parameters related to orientation of head estimation, specific facial landmarks, such as nose tip, chin, corners of the eye, and mouth corners are used. The process assumes no camera distortions during focal length estimation. The rotation vector \big\{$\mathbf{RV}_{[i,j]}^{3 \times 1} \big\}$, describing the object's orientation relative to the camera, is computed using the \texttt{solvePNP} function. 
\begin{eqnarray}
    {RV}_{[i,j]}^{3 \times 1} \xleftarrow{\text{solvePnP}}
\begin{pmatrix}
OP & : \text{Object Points} \\
IP & : \text{Image Points} \\
CM & : \text{Camera Matrix}
\end{pmatrix}
\end{eqnarray}
The \texttt{OpenCV} first extracts the axis `u' and angle $\theta$ from the rotation vector \big\{$\mathbf{RV}_{[i,j]}^{3 \times 1} \big\}$. It then constructs the skew-symmetric matrix K and calculates the R using the below formula. 
\begin{eqnarray}
R = I + \sin(\theta) \cdot K + (1 - \cos(\theta)) \cdot K^2
\end{eqnarray} 
The ${RV}_{[i,j]}^{3 \times 1}$  is converted to a rotation matrix 
${RM}_{[i,j]}^{3 \times 3}$ using the \texttt{Rodrigues} formula.   
 \begin{eqnarray}
{RV}_{[i,j]}^{3 \times 1}\xmapsto[R]{Rodrigues} {RM}_{[i,j]}^{3 \times 3}
\end{eqnarray} 
  
\noindent The ${RM}_{[i,j]}^{3 \times 3}$ is then used to extract the euler angles, as per the standard algorithm, to obtain the values of pitch $\theta_x$, yaw $\theta_y$, and roll $\theta_z$.
\begin{eqnarray}
R_{head}[r_{i,j}]_{3*3} \xleftarrow[Vector]{Euler}RM[\ ]_{N*N} \\
\theta_x \xleftarrow[Angle]{Euler}atan2(R_{head}\{r_{3,2},r_{3,3}\})\\
\theta_y \xleftarrow[Angle]{Euler}atan2(R_{head}\{-r_{3,1},[\sqrt{r^2_{3,3},r^2_{3,2}}]\})\\
\theta_z \xleftarrow[Angle]{Euler}atan2(R_{head}\{r_{2,1},r_{1,1}\})
\end{eqnarray} 

where R is the rotation matrix with elements $R_{ij}$. This concludes the  $\pounds_1^V$ of feature extraction thus obtaining EAR, pitch, yaw, and roll.

\subsubsection{Level $2$ ($\pounds_2^V$) Visual Feature Extraction }
 The next level, $\pounds_2^V$, focuses on deriving high-level categorical features, namely, eye gaze direction ($\varphi$), head position ($\hbar$), and eye openness category ($\epsilon$), from the numerical values obtained in $\pounds_1^V$. These categorical variables are computed from low-level geometric measurements based on facial landmarks to support robust and interpretable analysis of visual attention and head dynamics. By mapping raw measurements to meaningful categories, this stage captures semantically rich indicators of the user’s attentional focus and physiological state, particularly from the facial and ocular regions.
 

\vspace{.1in} \noindent \textbf{Gaze direction} represents the orientation of the user's line of sight relative to the camera. It is computed using a 2D gaze vector \(\vec{G}\), defined as the difference between the image center($\odot$) and the estimated eye center($\otimes$) :
\begin{eqnarray}
\vec{G} \xleftarrow{\odot-\otimes}\begin{cases}
\odot \xleftarrow{}\begin{cases}
    (\frac{Image\ Width}{2},\frac{Image\ Height}{2})
\end{cases} \\
\otimes\xleftarrow[/2]{\sum} \begin{cases}
\otimes_L = \frac{\sum_{i=36}^{i\leq41} p_{i}}{6}& \\
\otimes_R = \frac{\sum_{i=42}^{i\leq47} p_{i}}{6} & 
\end{cases}  

\end{cases} 
\end{eqnarray}

\begin{equation}
\vec{G} = \vec{C}_{\text{image}} - \vec{C}_{\text{eye}} = (G_x, G_y)
\end{equation}
\noindent where \(\vec{C}_{\text{image}} = (x_c, y_c)\) is the center of the image frame, and \(\vec{C}_{\text{eye}} = (x_e, y_e)\) is the detected center point of the eye region, $\otimes_L$ and $\otimes_R$ denotes center of left eye and right eye respectively. The components of \(\vec{G}\) reflect directional shifts in gaze:  
- \(G_x\) corresponds to horizontal motion (left/right)  
- \(G_y\) corresponds to vertical motion (up/down).

\noindent To translate the continuous values of the gaze vector into discrete gaze states, following threshold-based classification has been applied:

\begin{equation}
\varphi =
\begin{cases}
\text{Left} & \text{if } G_x > \tau_x^+ \\
\text{Right} & \text{if } G_x < \tau_x^- \\
\text{Up} & \text{if } G_y > \tau_y^+ \\
\text{Down} & \text{if } G_y < \tau_y^- \\
\text{Forward} & \text{if } \| G_x \| \leq \epsilon_x \text{ and } \| G_y \| \leq \epsilon_y
\end{cases}
\end{equation}

\noindent The thresholds \(\tau_x^{\pm}, \tau_y^{\pm}\) and neutral zone limits \(\epsilon_x, \epsilon_y\) were empirically determined based on pilot observations and calibrated for each subject. This discrete representation enables reliable detection of attentional shifts and off-screen gaze.

\vspace{.1in} \noindent \textbf{Head orientation} provides valuable cues about the user's attentiveness and focus. It is estimated from facial landmarks using a perspective-n-point (PnP) solution, such as OpenCV's \texttt{solvePnP} function, which outputs a rotation vector. This vector is decomposed into three Euler angles: pitch (\(\theta_x\)), yaw (\(\theta_y\)), and roll (\(\theta_z\)). The following are the seven head position classes based on these angles:

\begin{equation}
\hbar =
\begin{cases}
\text{Neutral} & \text{if } \, \lvert \theta_x \rvert,\, \lvert \theta_y \rvert,\, \lvert \theta_z \rvert \leq \epsilon_\theta, \\
\text{Tilted Up} & \text{if } \theta_x > \tau_x^{+}, \\
\text{Tilted Down} & \text{if } \theta_x < \tau_x^{-}, \\
\text{Turned Left} & \text{if } \theta_y > \tau_y^{+}, \\
\text{Turned Right} & \text{if } \theta_y < \tau_y^{-}, \\
\text{Tilted Left} & \text{if } \theta_z > \tau_z^{+}, \\
\text{Tilted Right} & \text{if } \theta_z < \tau_z^{-}.
\end{cases}
\end{equation}

\noindent Here, \(\epsilon_\theta\) defines the margin for neutral alignment, while \(\tau_x^\pm, \tau_y^\pm, \tau_z^\pm\) define directional tilt/turn thresholds. This classification helps distinguish between engaged (neutral/forward-facing) and disengaged (turned/tilted) head positions.

\vspace{.1in} \noindent The \textbf{degree of eye openness}  $\epsilon$ is an important physiological indicator for detecting blinking, fatigue, or drowsiness. We use EAR, denoted by \(\xi\), to quantify the vertical-to-horizontal ratio of eye landmark distances:

\begin{equation}
\xi = \frac{\|p_2 - p_6\| + \|p_3 - p_5\|}{2 \cdot \|p_1 - p_4\|}
\end{equation}

\noindent where \(p_1\) through \(p_6\) are six annotated landmarks on the eye contour. Based on EAR, eye openness \(\epsilon\) is categorized into three discrete states:

\begin{equation}
\epsilon =
\begin{cases}
\text{Fully Open} & \text{if } \xi \geq T_1 \\
\text{Partially Closed} & \text{if } T_2 < \xi < T_1 \\
\text{Closed} & \text{if } \xi \leq T_2
\end{cases}
\end{equation}
\noindent The thresholds \(T_1\) and \(T_2\) are determined through initial calibration or dataset statistics. This feature plays a key role in fatigue and attention monitoring applications.

The final set of categorical features \(\{\varphi, \hbar, \epsilon\}\) serves as high-level descriptors of user attention and engagement. These features are computed using interpretable geometric transformations and threshold-based rules, making them reliable for downstream behavior recognition models. The discrete nature of these variables enables robust classification, especially in noisy or real-time environments.

In contrast to deep feature-based approaches, this geometry-driven, interpretable extraction strategy enables low-latency inference, transparent decision-making, and resilience to computational constraints. These properties make it particularly suited for real-time educational applications where both interpretability and responsiveness are essential.

\subsection{Physiological Feature Extraction Phase} \label{phy}
The second phase of feature extraction targets the estimation of physiological signals, specifically the Blood Volume Pulse (BVP), using the POS method for rPPG. This non-contact technique relies on subtle skin color variations in facial video frames to capture cardiovascular activity. In Level 1 ($\pounds_1^P$), raw RGB signals are extracted from a stable forehead region of interest (ROI) using facial landmarks. The signals are normalized, projected using the POS transformation, and refined through detrending, illumination correction, and temporal filtering to enhance the BVP signal quality. In Level 2 ($\pounds_2^P$), the processed BVP signal is analyzed to extract higher-level physiological features such as RR intervals, average heart rate (BPM), peak-to-peak intervals (PPI), and cumulative systolic/diastolic peaks. These features provide insight into cardiovascular dynamics and user physiological state. The subsequent sections provide a detailed explanation of each level.

\subsubsection{Level 1 ($\pounds_1^P$) Physiological Features }
The $\pounds_1^P$ denotes the first level in extracting physiological features, which begins with extracting signals ($S$) by calculating the average RGB values from a specified ROI i.e. forehead $(\Re_f)$ for the proposed model in each video frame.  A consistent and reliable ROI must be identified on the face. This is achieved using a face recognition library to detect facial landmarks, particularly around the eyebrows and forehead. The forehead is selected as the ROI due to its minimal movement and lack of hair, making it an ideal area for capturing subtle colour changes indicating blood flow.

\begin{equation}
\mathbf{S}_{t} = \frac{1}{N} \sum_{i=1}^{N} 
\begin{bmatrix}
R_{i}(t) \\
G_{i}(t) \\
B_{i}(t)
\end{bmatrix}
\end{equation}
where $R_i(t), G_i(t),$ and $B_i(t)$ are the RGB values from the region of interest in the $i^{th}$ video frame, and N is the total number of pixels in the $(\Re_f)$. 

To ensure the quality of rPPG signals during feature extraction, a series of control measures were applied to minimize the impact of measurement noise, motion artifacts, and illumination changes. These steps significantly reduced spectral leakage and variability, resulting in clean and reliable BVP signals suitable for accurate downstream physiological feature computation.
\begin{itemize}
    \item  Stable forehead ROI tracking with landmark checks.
    \item  Rejection of frames with occlusion, excessive inter–frame motion ($>$8 px) or $<$80\% skin pixels (YCbCr mask).
    \item Illumination normalization (per‑channel mean division, detrending) and interpolation of intensity outliers ($|z|>$3).
    \item POS selection over GREEN after higher window SNR (replaceable with exact values).
    \item 4th‑order zero‑phase Butterworth band‑pass (0.7–4 Hz)
    \item Sliding 200‑frame windows discarded if dominant spectral peak absent or energy ratio (physio band / total) $<$0.4.
    \item Adaptive peak detection with refractory $\geq$0.3 s, removal or interpolation of implausible IBIs outside 0.3–1.5s. 
    
\end{itemize}

\noindent Further, temporal normalization is applied to the RGB values by dividing each colour signal by its mean over a specific time window. This transformation emphasizes the pulsatile component, primarily captured in the green channel, which is most sensitive to blood flow changes. The normalized signals are calculated as:
\begin{equation}
    R_{\text{norm}}(t) = \frac{R(t)}{\mu_R}, \quad G_{\text{norm}}(t) = \frac{G(t)}{\mu_G}, \quad B_{\text{norm}}(t) = \frac{B(t)}{\mu_B}
\end{equation}
where \( \mu_R \), \( \mu_G \), and \( \mu_B \) are the mean values of the \( R(t) \), \( G(t) \), and \( B(t) \) signals over the time window, respectively. This process reduces noise and variations unrelated to the pulse signal, ensuring the output reflects pulsatile changes due to blood flow.

\noindent Following normalization, the signal is projected onto a plane orthogonal to the skin tone vector using a transformation matrix. This matrix emphasizes the pulsatile component of the signal, primarily captured in the green channel, which is most sensitive to changes in blood flow. Following normalization, the signal is projected onto a plane orthogonal to the skin tone vector using a transformation matrix $(T)$. The projection is defined as:
\begin{eqnarray}
    \mathbf{S}_{\text{proj}}(t) = \mathbf{T} \cdot \mathbf{S}_{\text{norm}}(t) \\ \nonumber 
      \mathbf{S}_{\text{norm}}(t) = \begin{bmatrix} R_{\text{norm}}(t) \\ G_{\text{norm}}(t) \\ B_{\text{norm}}(t) \end{bmatrix}
\end{eqnarray}
This transformation emphasizes the pulsatile component, primarily captured in the green channel, which is most sensitive to changes in blood flow.

Further refinement is performed by tuning the signal components based on their standard deviation, enhancing the accuracy of the BVP signal by minimizing noise and artifacts, such as motion or lighting changes. In the next step, refinement is performed by tuning the signal components based on their standard deviation ($\sigma_S $). This step enhances the accuracy of the BVP signal by minimizing noise and artifacts, such as motion or lighting changes. The refined signals are calculated as: 

\begin{equation}
\mathbf{S}_{\text{refined}}(t) = \frac{\mathbf{S}_{\text{proj}}(t)}{\sigma_S}
\end{equation}

 \noindent The POS algorithm is then applied to the RGB values extracted from this region across video frames, enhancing the BVP signal's accuracy and reliability.
To effectively apply the POS algorithm, we first need to identify the ROI \textit{i.e.} ($\Re_f$).
\begin{eqnarray}
     \Re_f= FLM((p_1) - (p_{68}))\\ \nonumber
     \text{RGB}(t) = \text{ExtractRGB}(\Re_f, t)
\end{eqnarray}
Finally, to extract the BVP ($\mathbf{B}$) signal, the POS algorithm has been applied to the extracted $\Re_f$.  
\begin{equation}
\text{B}(t) = POS_{Algo}(\text{RGB}(t))
\end{equation}

The signal is first detrended to eliminate any long-term trends that might obscure the true pulse. A band-pass filter is then applied to isolate the frequency components corresponding to the heart rate. The filtered signal is analyzed to detect peaks representing heartbeats, and the intervals between these peaks, known as RR intervals, are calculated to determine the heart rate in BPM.

\subsubsection{Level 2 ($\pounds_2^P$) Physiological Features }
After obtaining the BVP signal from $\pounds_1^P$, further in $\pounds_2^P$ analysis is conducted to extract more physiological features. Then the BVP signal is detrended to remove long-term trends and, a band-pass filter is applied to isolate the heart rate frequency components from the $\tilde{B}(t)$. 
\begin{equation}
    \tilde{B}(t) = B(t) - \text{Trend}(B(t)) \ \ || \quad B_{\text{filtered}}(t) = \mathcal{F}_{\text{BP}}(\tilde{B}(t))  
\end{equation}
where \( B(t) \) is the original BVP signal, and \( \tilde{B}(t) \) is the detrended signal.
\begin{equation}
B_{\text{filtered}}(t) = \mathcal{F}_{\text{BP}}(\tilde{B}(t))    
\end{equation}
\( \mathcal{F}_{\text{BP}} \) denotes the band-pass filter, and \( B_{\text{filtered}}(t) \) is the filtered signal. To calculate the $RR_i$ interval, analysis has been done to identify peaks representing heartbeats $(t_{i+1}-t_i)$. 
\begin{equation}
\forall_{i=1}^{i\leq N} \big\{RR_i = t_{i+1} - t_i\big\} 
\end{equation}
where \( t_i \) are the timestamps of the detected peaks and \( RR_i \) are the RR intervals. 
From the mean value \( \overline{RR} \), heart rate in BPM is given by: 
\begin{equation}
\text{BPM} = \frac{\overline{RR}}{60}   
\end{equation}  
such as the average peak-to-peak interval, are extracted to gain insights into heart rhythm regularity. The cumulative sums of systolic and diastolic peaks are also computed to provide a comprehensive overview of cardiovascular activity. The entire process culminates in the computation of the average heart rate, summarizing the overall trend of the heart rate based on the captured BVP signal. This robust approach ensures accurate and reliable heart rate estimation from video frames, demonstrating the efficacy of the POS method for remote photoplethysmography.
Some more features in $\pounds_2^P$, like average $PPI$ to assess heart rhythm regularity, are calculated as follows:  
\begin{equation}
    \text{PPI}_{\text{avg}} = \frac{1}{N} \sum_{i=1}^{N} RR_i
\end{equation} 
where \( RR_i \) is the interval between consecutive peaks, and \( N \) is the total number of detected RR intervals.
Similarly, from $RR_i$, the cumulative sums of systolic $(CS_{Sys})$ and diastolic $(CS_{Dia})$ peaks are extracted.
\begin{equation}
CS_{Sys} = \sum_{i=1}^{N} S_i \quad \text{and} \quad CS_{Dia} = \sum_{i=1}^{N} D_i
\end{equation}
where \( S_i \) and \( D_i \) represent the systolic and diastolic peaks, respectively.
The average heart rate over the entire signal duration is computed, and overall heart rate trends are summarized based on the BVP signal.
\begin{eqnarray}
    \text{HR}_{\text{trend}} = \frac{1}{T} \int_0^T \text{BPM}(t) \, dt
\end{eqnarray}where \( \overline{RR} \) is the mean RR interval over the entire recording period. \( \text{BPM}(t) \) is the instantaneous heart rate at time \( t \), and \( T \) is the total duration of the signal. 

%
%

By incorporating a non-contact rPPG POS method into the pipeline, VisioPhysioENet extends physiological monitoring capabilities to standard webcams without requiring specialized sensors. This integration represents one of the first such applications in multimodal engagement detection, enabling scalable and cost-effective deployment.

\subsection{Multimodal Fusion Phase} 
In our proposed architecture, we implement a Late Fusion strategy to effectively integrate Visual and Physiological features for enhanced classification performance. The visual features extracted in section \ref{vis} $\pounds_2^V$ are processed using \texttt{AdaBoost} Classifier, to compute the probability distribution, $P_{AB}(V)$.   
\begin{eqnarray}
P_{\text{AB}}(V) = \text{AdaBoost}(\{ \varphi, \hbar, \epsilon \} ) 
\end{eqnarray} 
where \( P_{\text{AdaBoost}}(V) \) is the probability distribution output from the AdaBoost classifier based on these visual features. This approach ensures that the nuances in visual cues are accurately captured and leveraged for the classification task.

The Visual features, which include `eye category', `eye position', `gaze direction', `ear', `pitch', `yaw', and `roll', are critical in capturing intricate facial expressions and gaze patterns. These features are processed using an AdaBoost Classifier, which is particularly effective in boosting the accuracy of weak learners by combining them into a strong ensemble model. 

On the other hand, the Physiological features from $\pounds_2^P$ computed in section \ref{phy} are used for assessing the subject's physiological responses using a \texttt{Random Forest} Classifier. 
\begin{eqnarray}
P_{\text{RF}}(P) = RandFor\big\{ PPI_{avg},CS_{Sys}, CS_{Dia}, HR_{trend}) \big\}
\end{eqnarray}

The physiological features, comprising `heart rates', `peak-to-peak intervals', `systolic peaks' and `diastolic peaks', are vital for assessing the subject's physiological responses. These features are handled by a Random Forest Classifier, known for its robustness in handling high-dimensional and complex data. The Random Forest model generates probability distributions for the Physiological features, reflecting the likelihood of various classification outcomes.\\

\noindent 
\textbf{Late Fusion via Stacked Generalization}:
In the final stage of our architecture, a Late Fusion strategy is implemented to combine the outputs of the visual and physiological feature classifiers. The visual features, denoted \( V \), are processed using an AdaBoost classifier, while the physiological features, denoted \( P \), are classified using a Random Forest model. These two classifiers output probability distributions over class labels, expressed as:

\begin{equation}
P_{\text{AB}}(V) = \text{AdaBoost}(\{\varphi, \hbar, \epsilon\})
\end{equation}

\begin{equation}
P_{\text{RF}}(P) = \text{RandomForest}(\{ \text{PPI}_{\text{avg}}, CS_{\text{Sys}}, CS_{\text{Dia}}, \text{HR}_{\text{trend}} \})
\end{equation}

To integrate these modality-specific predictions, we apply a stacked generalization meta-classifier (e.g., logistic regression), which takes both probability vectors as input and computes the final prediction probability:

\begin{equation}
P_{\text{final}} = \text{MetaClassifier}\big(P_{\text{AB}}(V), P_{\text{RF}}(P)\big)
\end{equation}

The final predicted class label \( \hat{y} \) is selected by applying the \(\arg\max\) operator over the final probability distribution:

\begin{equation}
\hat{y} = \arg\max_y \Big\{ P_{\text{final}}(y) \Big\}
\end{equation}

This fusion mechanism leverages the complementary strengths of AdaBoost (which captures subtle visual dynamics) and Random Forest (which robustly handles physiological variations) to form a more accurate and reliable classification model. By integrating multi-modal data at the decision level, the system achieves improved generalization and prediction robustness across varying subjects and conditions.

\section{Experimental Results}
\subsection{Implementation Setup}\label{sec:setup}
All model training and evaluation were carried out on a workstation equipped with an Intel(R) Core(TM) i7-7700 CPU operating at 3.70 GHz and an NVIDIA RTX 2070 GPU. Implementation of both the neural machine translation and sentiment analysis models was performed using the PyTorch\footnote{\url{https://pytorch.org/}} library, leveraging GPU acceleration to ensure efficient computation throughout training and inference.

\subsection{Dataset, Training Strategy and Hyper-parameters}\label{sec:dataset}
The experiments presented in this study exclusively utilize the DAiSEE dataset \cite{gupta2016daisee}, a comprehensive benchmark specifically designed for analyzing learner engagement in educational environments. The DAiSEE dataset comprises a total of 9,068 video recordings collected from 112 adult participants. Each video has been meticulously annotated by expert human annotators into engagement levels, providing a robust foundation for supervised learning tasks. For training and evaluation purposes, we adopted an 80-20 train-test split strategy with 10-fold cross-validation. The visual stream was sampled at 5\,fps, with EAR thresholds of 0.28 (open) and 0.20 (closed), neutral gaze defined within $\pm 6$\,px, and neutral head pose within $\pm 8^{\circ}$ (turn/tilt thresholds $\pm 12^{\circ}$). rPPG extraction used 200-frame sliding windows (50\% overlap) band-pass filtered at 0.7--4\,Hz, with low-quality windows discarded based on signal-quality gating. The visual branch employed an AdaBoost classifier with 150 depth-1 estimators (learning rate = 0.5, balanced class weights), while the physiological branch used a Random Forest with 300 trees (max depth = 12, min samples split = 4, balanced class weights). Outputs from both branches were fused via logistic regression (L2 penalty, $C=1.0$, balanced class weights).

\subsection{Models}\label{sec:models}
To rigorously evaluate the efficacy of our proposed VisioPhysioENet, we established three baseline models alongside our final proposed configuration:\vspace{.04in}

\begin{itemize}
\item \textbf{Baseline 1 (Visual Modality Only)}: This configuration exclusively utilizes visual cues extracted from facial landmarks, head positioning, and gaze direction. Multiple classifiers were employed, including AdaBoost, Voting Classifier, Decision Tree, Random Forest, Gradient Boosting, Support Vector Classifier, and K-Nearest Neighbors, to evaluate their performance. \vspace{.04in}

\item \textbf{Baseline 2 (Physiological Modality Only)}: This configuration leverages physiological signals derived from heart rate variability, peak-to-peak intervals, and cumulative sums of systolic and diastolic peaks. The same set of classifiers as in Baseline 1 was applied to assess physiological data independently.\vspace{.04in}

\item \textbf{Baseline 3 (Early Fusion of Visual and Physiological Modalities)}: Visual and physiological modalities were combined through simple early fusion, concatenating features before classification. Classifiers identical to those in Baselines 1 and 2 were utilized for consistency and comparative analysis.\vspace{.04in}

\item \textbf{VisioPhysioENet (Late Fusion of Visual and Physiological Modalities)}: The proposed configuration employs a sophisticated late fusion approach. It first processes visual features using an AdaBoost classifier and physiological features using a Random Forest classifier. Their probability outputs are subsequently integrated via stacked generalization with a logistic regression meta-classifier to provide final engagement predictions.
\end{itemize}

\subsection{Results}\label{sec:results}

\subsubsection{Quantitative Results}\label{sec:results}
VisioPhysioENet achieves an overall accuracy of 63.09\% on the DAiSEE dataset. We first benchmark this performance against existing state-of-the-art methods to assess its competitiveness in multimodal engagement detection, followed by a comparison with baseline models to quantify the gains from each design choice. \\

\noindent \textbf{Comparison with the state-of-the-art}: 
Table \ref{tab:sota} compares VisioPhysioENet's performance with the state-of-the-art methods. Notably, it outperforms the existing methods including the ``Unsupervised Behavior \& rPPG" model by Vedernikov et al. \cite{vedernikov2024analyzing}, which also combines these modalities but with less efficacy. The proposed system achieves an accuracy improvement of 8.6\% over this model, demonstrating the effectiveness of the late fusion approach in handling multimodal data. This significant improvement in performance highlights VisioPhysioENet's potential to advance multimodal engagement detection, setting a new benchmark for future studies.

\begin{table}[!h]
\centering
{\selectfont
    \caption{Results' comparison with state-of-the-art. Here, `V,' `P' and `Acc' denote visual \& physiological modalities and accuracy.} 
    \label{tab:sota}
    \resizebox{.72\textwidth}{!}
    {%
    \begin{tabular}{lcc}\toprule
        \textbf{Model} & \textbf{Modality} & \textbf{Acc}\\ \midrule
        InceptionNet Video Level \cite{gupta2016daisee} & V & 46.40\%\\ 
        InceptionNet Frame Level \cite{gupta2016daisee} &  V & 47.10\%\\
        Class-Balanced 3D CNN \cite{zhang2019annovel} & V & 52.35\%\\ 
        ResNet +  Temporal Conv Net \cite{abedi2021improving} & V & 53.70\%\\  
        Unsupervised Behavior \& rPPG \cite{vedernikov2024analyzing} & V+P & 54.49\%\\ 
        Long-Term Recurrent Conv Net \cite{gupta2016daisee} & V & 57.90\%\\
        Deep Spatiotemporal Net \cite{liao2021deep} & V & 58.84\%\\
        C3D + TCN \cite{abedi2021improving} & V & 59.97\%\\
        ResNet + LSTM\cite{abedi2021improving} & V & 61.15\% \\ 
        \textbf{VisioPhysioENet}  & \textbf{V+P} & \textbf{63.09}\% \\ 
        \bottomrule
    \end{tabular}%
    }
}
\end{table}

Despite not achieving the highest accuracy, VisioPhysioENet stands out by leveraging both physiological and visual data, providing a richer, more nuanced understanding of engagement. It outperforms the only other multimodal model by 8.6\%, demonstrating its superior ability to integrate these data types. The unique features of VisioPhysioENet, such as advanced feature extraction and fusion techniques, enhance its utility in real-world educational settings. Additionally, the robustness and generalizability  of the model suggest it can effectively handle variations and imbalances across different datasets, making it a valuable tool for practical applications.\\

\noindent \textbf{Comparison with baseline models}: 
While the state-of-the-art comparison demonstrates competitiveness, the baseline evaluation isolates the contribution of each modality and fusion strategy. Table \ref{tab:sota_bl} positions VisioPhysioENet against the baseline models described in Section \ref{sec:models}. 

\begin{table}[!h]
\centering 
    \caption{Results' comparison with the baseline models. Here, `V' and `P' represent Visual and Physiological modalities.} 
    \label{tab:sota_bl}
    \resizebox{.65\textwidth}{!}
    {%
    \begin{tabular}{@{}lccc}\toprule
        \textbf{Model} & \textbf{Modality} & \textbf{Fusion Strategy} & \textbf{Acc} \\ \midrule
        Baseline 1 & V  & \textendash & 43.03\% \\
        Baseline 2 & P  & \textendash & 55.25\% \\
        Baseline 3 & V+P & Early Fusion & 51.50\% \\ 
        \textbf{Proposed}   & \textbf{V+P} & \textbf{Late Fusion} & \textbf{63.09}\% \\ \bottomrule
    \end{tabular}%
    }
\end{table}
 
The results reveal three main observations: (i) physiological cues alone (55.25\%) surpass the visual-only stream (43.03\%), underlining the discriminative power of micro-vascular dynamics; (ii) a naive feature-level (early) fusion actually degrades performance (51.50\%), suggesting incompatible feature spaces when modalities are treated homogeneously; (iii) our late-fusion strategy, which allows modality-specific classifiers to specialise before stacking, delivers 63.09\% with a gain of +7.8\% over the best baseline and +20.1\% over the visual-only model. It validates the core design choice of VisioPhysioENet.

\subsubsection{Qualitative Results}\label{sec:qualitative}
\begin{figure}[!h]
\centering
\includegraphics[width=1\textwidth]{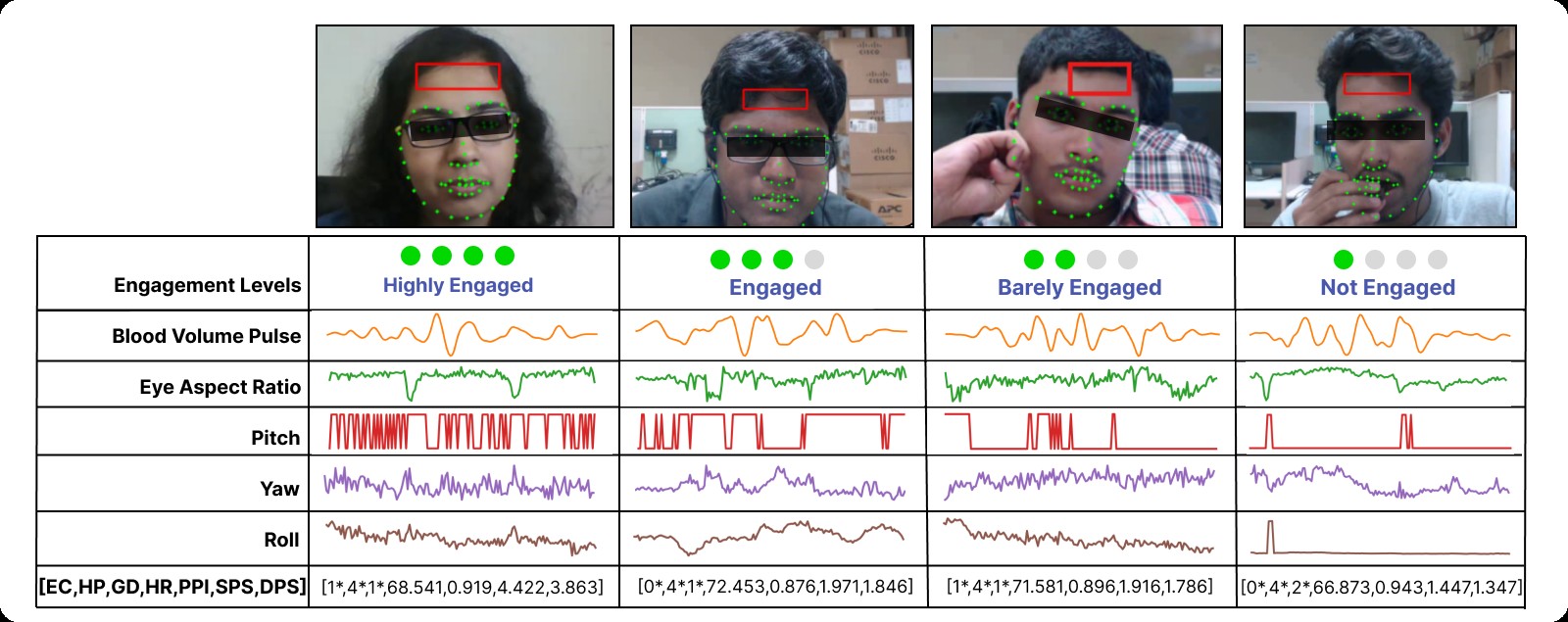} 
\caption{Sample Results. Here, EC: Eye Category, HP: Head Position, GD: Gaze Direction, HR: Heart Rate, PPI: Peak-to-Peak Interval, SPS: Systolic Peak Sum, DPS: Diastolic Peak Sum; $^*$ represents categorical values while others are numeric values.}  
\label{fig:qual_res}
\end{figure}

Fig.~\ref{fig:qual_res} presents representative qualitative outputs generated by the proposed VisioPhysioENet system for four distinct engagement levels: Highly Engaged, Engaged, Barely Engaged, and Not Engaged. The visual modality features include head position (quantified using pitch, yaw, and roll), gaze direction (derived from 2D eye-center displacement relative to the image center), and eye openness category (computed from EAR). The physiological modality features are extracted from the BVP signal and include heart rate (HR), PPI, systolic peak sum (SPS), and diastolic peak sum (DPS). In the plotted BVP traces, systolic peaks correspond to crests while diastolic peaks align with troughs.

By simultaneously analyzing visual and physiological cues, VisioPhysioENet effectively captures both overt and subtle behavioral signatures indicative of engagement. For instance, a \emph{Highly Engaged} learner maintains a steady gaze toward the screen, stable head orientation, consistent EAR values, and a regular cardiac rhythm with uniform PPI intervals. An \emph{Engaged} learner exhibits slightly more variability in gaze and head position while still retaining relatively stable physiological signals. In contrast, \emph{Barely Engaged} learners display intermittent gaze aversion, more frequent head movements, and increased variability in EAR and PPI values. The \emph{Not Engaged} category is characterized by significant deviations in gaze direction, abrupt head posture changes, irregular eye openness, and highly fluctuating cardiovascular indicators. These multimodal patterns highlight the system’s ability to differentiate engagement levels with fine-grained precision by leveraging complementary visual and physiological evidence.

\subsection{Ablation Studies}\label{sec:ablation}
To further validate and optimize the proposed system, several ablation studies were conducted, each examining critical choices in feature extraction and fusion methodologies.  

\subsubsection{Choice of Appropriate Fusion Approach} 
To identify the most effective strategy for integrating visual and physiological modalities, we evaluated multiple fusion approaches, including unimodal baselines, early fusion, and our proposed late fusion via stacked generalization. In the proposed configuration, AdaBoost focuses on iteratively correcting misclassifications, while Random Forest leverages ensemble averaging across decision trees to enhance robustness. A logistic regression meta-classifier then combines their probability outputs, capturing complementary patterns and improving predictive performance. This late fusion design achieves an accuracy of 63.09\%, surpassing all baseline configurations.

\begin{figure}[!h]
    \centering
    \subfloat[Visual-only Accuracy Results\label{fig:enter-label}]{%
        \includegraphics[width=0.62\linewidth]{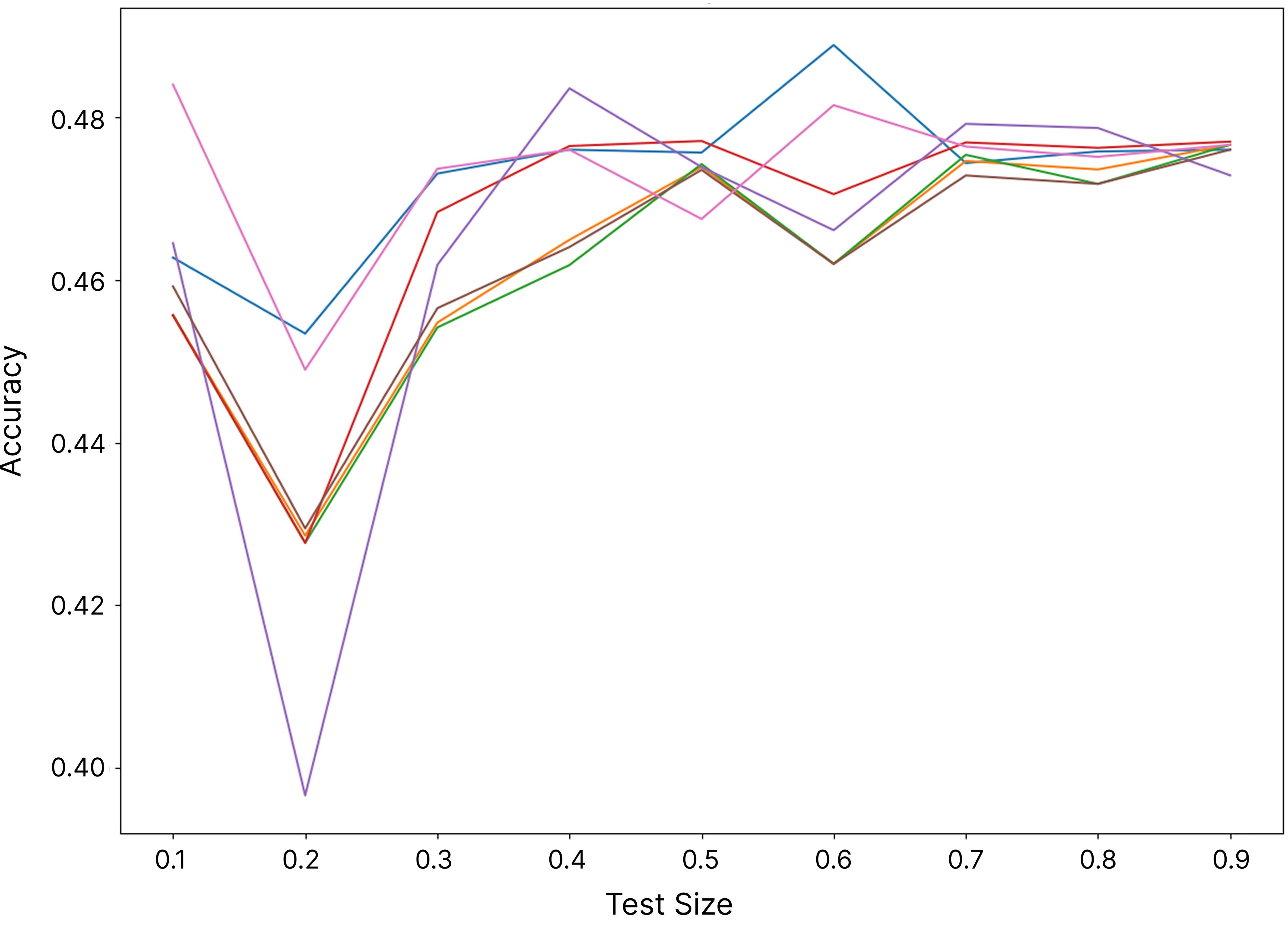}
    }\vspace{-.1in}

    \subfloat[Physiological-only Accuracy Results\label{fig:res_P_only}]{%
        \includegraphics[width=0.62\linewidth]{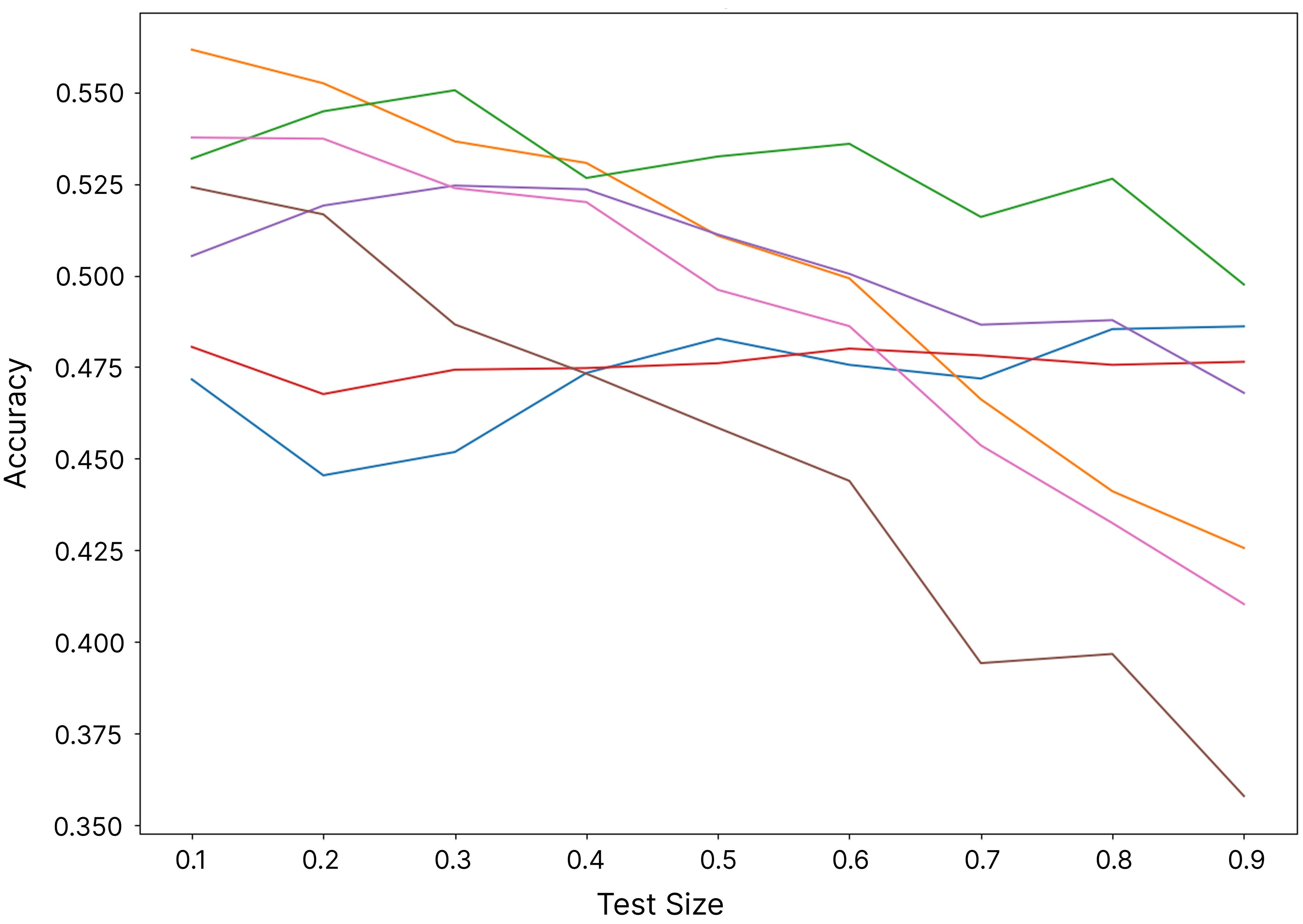}
    }\vspace{-.1in}

    \subfloat[Early Fusion (V+P) Accuracy Results\label{fig:enter-label2}]{%
        \includegraphics[width=0.62\linewidth]{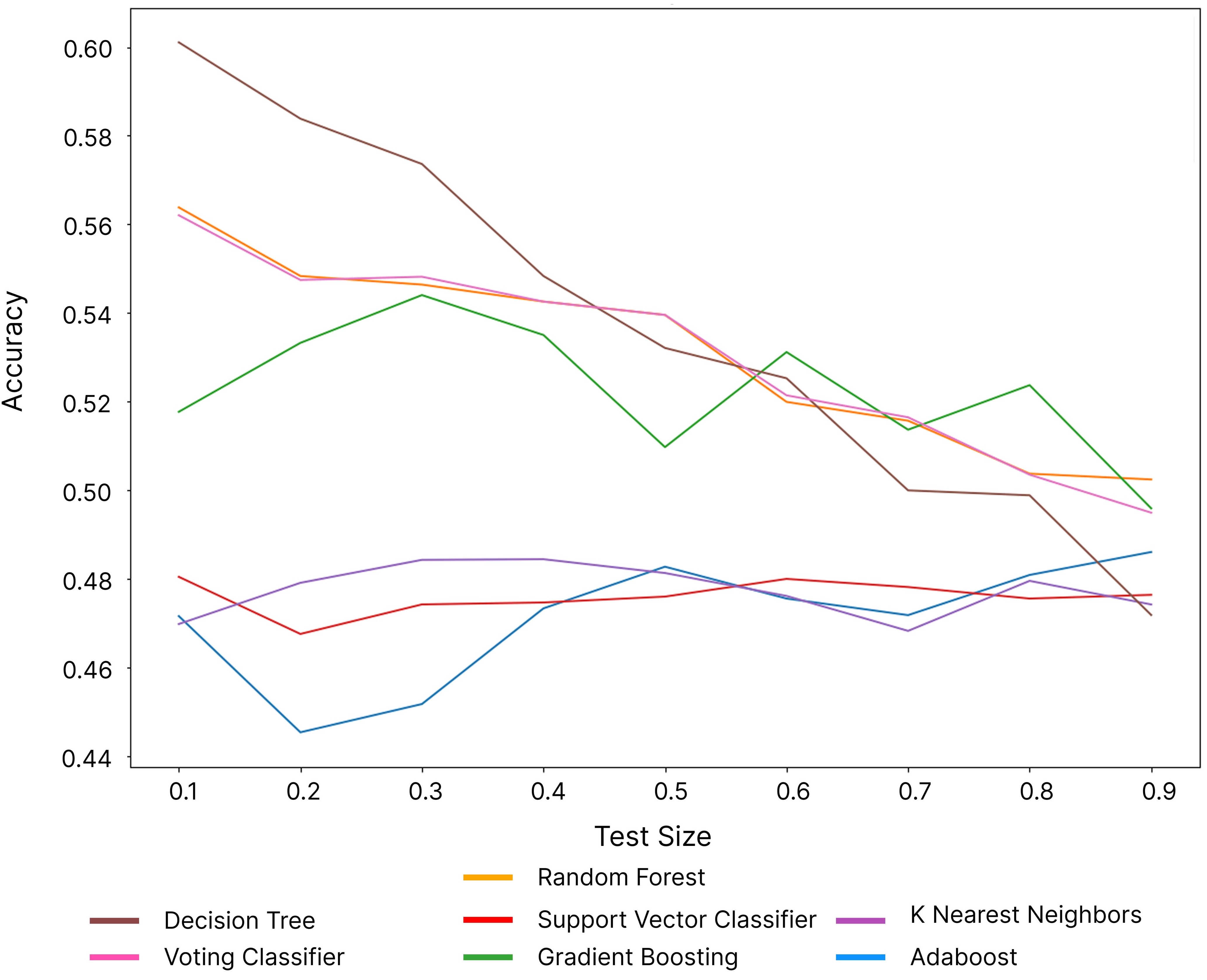}
    }

    \caption{Unimodal and early fusion accuracy results}
    \label{fig:baseline_results_combined}
\end{figure}

As depicted in Fig. \ref{fig:baseline_results_combined}, Baseline 1 (visual-only) showed variation in accuracy across classifiers, with AdaBoost achieving the highest performance at 45.34\%, followed by Voting Classifier (44.90\%). Decision Tree reached 42.95\%, while Random Forest and Gradient Boosting both scored 42.80\%. The Support Vector Classifier matched Gradient Boosting, and K-Nearest Neighbors (KNN) was the least accurate at 39.66\%, yielding an overall average accuracy of 43.03\%. Baseline 2 (physiological-only) outperformed the visual-only baselines, with Random Forest leading at 63.27\%, followed by Decision Tree (63.09\%) and Voting Classifier (62.73\%). Gradient Boosting and KNN achieved 54.48\% and 51.91\%, respectively, while AdaBoost and Support Vector Classifier recorded 44.54\% and 46.76\%, giving an overall mean accuracy of 55.25\%. Baseline 3 (early fusion of visual and physiological features) demonstrated intermediate performance, with Decision Tree achieving 58.39\%, followed by Random Forest (54.84\%) and Voting Classifier (54.75\%). Gradient Boosting reached 53.33\%, KNN scored 47.91\%, and AdaBoost and Support Vector Classifier recorded the lowest accuracies at 44.54\% and 46.76\%, resulting in an overall average accuracy of 51.50\%.

Overall, VisioPhysioENet’s stacked generalization late fusion delivers a +7.8\% gain over the best baseline and +20.1\% over the visual-only model, confirming its effectiveness in leveraging complementary visual and physiological cues for robust learner engagement prediction.

\subsubsection{Choice of Machine Learning Models}
Tables \ref{tab:mlvisual} and \ref{tab:mlphysio} outline the machine learning models selected for visual and physiological modalities. The AdaBoost Classifier, achieving an accuracy of 43.03\% for visual features, is chosen for its ability to enhance performance by focusing on difficult cases in complex visual data, such as variations in lighting and expression. In contrast, the Random Forest Classifier, which attains a higher accuracy of 55.25\% for physiological features, is utilized due to its robustness in handling multivariate data and the feature selection capabilities essential for physiological signals such as heart rate and blood pressure peaks. 

\begin{table}[!h]
\centering
{\selectfont
    \caption{Choice of ML Models for Visual Modality.} 
    \label{tab:mlvisual}
    \resizebox{.52\textwidth}{!}
    {%
    \begin{tabular}{@{}lcccc}\toprule
        \textbf{Model} & \textbf{Accuracy} \\ \midrule
        K-Nearest Neighbors Classifier & 39.66\% \\
        Gradient Boost Classifier & 42.77\%   \\
        Support Vector Classifier & 42.77\% \\
        Random Forest Classifier & 42.86\% \\
        Decision Tree Classifier & 42.95\% \\
        \textbf{Ada Boost Classifier} & \textbf{45.03}\%\\ \bottomrule
    \end{tabular}%
    }
}
\end{table}

\begin{table}[!h]
\centering
{\selectfont
    \caption{Choice of ML Models for Physiological Modality.} 
    \label{tab:mlphysio}
    \resizebox{.52\textwidth}{!}
    {%
    \begin{tabular}{@{}lcccc}\toprule
        \textbf{Model} & \textbf{Accuracy} \\ \midrule
        AdaBoost Classifier & 44.54\%\\
        Support Vector Classifier & 46.76\%\\
        Decision Tree Classifier & 51.67\%\\
        K-Nearest Neighbors Classifier & 51.91\%\\
        Gradient Boost Classifier & 54.48\%\\
        \textbf{Random Forest} & \textbf{55.25}\%\\ \bottomrule
    \end{tabular}%
    }
}
\end{table}

\subsubsection{Choice of Appropriate Visual Features}
To identify the most suitable visual features for effective engagement detection, we explored different approaches for categorizing continuous facial metrics into meaningful engagement indicators. Two strategies were evaluated: manually defined threshold-based categorization and automatic, data-driven feature extraction without explicit thresholding. The manual approach involved setting specific thresholds based on domain knowledge and empirical analysis for EAR, head pose (pitch, yaw, roll), and gaze direction, enabling the conversion of continuous metrics into categorical variables. This approach provided interpretability and simplicity but risked limited generalizability across diverse datasets. Conversely, the data-driven method leveraged raw, continuous visual features directly, allowing the machine learning models to autonomously detect subtle patterns without relying on explicit thresholds. While this provided greater flexibility and adaptability, it reduced interpretability. After experimentation, the threshold-based categorical features offered a balanced compromise between accuracy, interpretability, and robustness, proving more suitable for reliable engagement classification across varied scenarios.

\subsubsection {Choice of Appropriate Physiological Features}
Two techniques were used to extract rPPG signals, which measure changes in blood volume through the skin. The first method, GREEN, and the second, POS. The GREEN method produced signals with a lower signal-to-noise ratio (SNR), meaning the quality of the signals was less reliable due to higher levels of noise. In contrast, the POS method delivered clearer signals with a higher SNR, making it the preferred choice for this study. For selecting the regions of interest (ROIs) on the face where the rPPG signals were extracted, three options were considered: the forehead, left cheek, and right cheek. These areas were chosen based on their ability to closely match the true PPG signal. However, the forehead was chosen as the final ROI for this paper. This decision was made because the left and right cheeks could be difficult to use for accurate signal extraction in individuals with beards, as facial hair can obstruct or distort the measurements.

\subsubsection{Computational Runtime Analysis}\label{sec:ablation4}
Pre-extracting relevant features from raw video frames allows for parallel processing, which reduces overall system latency and optimizes CPU utilization. Sequential task execution often results in idle processor time, as each task waits for the previous one to complete. To address this, we implemented the ThreadPoolExecutor \cite{newvirtual}, enabling concurrent execution of multiple tasks across different threads. This approach minimizes idle time, enhances throughput, and makes better use of multi-core CPU resources. Table \ref{tab:comptime} reports the execution times for sequential and parallel methods during the processing of a single batch of varying sizes. Parallel processing delivers consistent speed improvements, with greater gains for larger batch sizes, as also illustrated in Fig.~\ref{fig:enter-label}.

\begin{table}[!h]
\centering
\caption{Comparison of execution times for sequential (ST) and parallel (TPE) methods using ThreadPoolExecutor.}
\label{tab:comptime}
\resizebox{.46\textwidth}{!}{%
\begin{tabular}{ccc} \toprule
\textbf{Batch Size} & \textbf{ST (sec)} & \textbf{TPE (sec)} \\
\midrule
1  &  7.4   &  7.0   \\
2  & 14.1   &  8.8   \\
4  & 27.2   & 12.4   \\
8  & 55.8   & 21.9   \\
16 & 109.7  & 44.9   \\
32 & 255.6  & 100.0  \\
64 & 732.6  & 293.0  \\ \bottomrule
\end{tabular}%
}
\end{table}

\begin{figure}[!h]
    \centering
    \includegraphics[width=.85\linewidth]{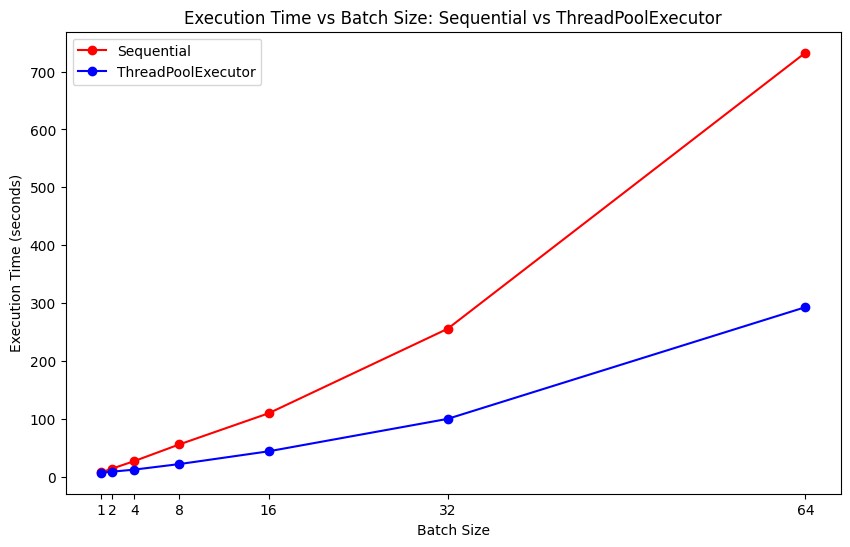}
    \caption{Computational runtime analysis using ThreadpoolExecuter}
    \label{fig:enter-label}
\end{figure}

These results confirm that VisioPhysioENet maintains low computational overhead, with a parallelized feature extraction pipeline significantly reducing processing times even for large batches. Importantly, the framework achieves these gains on standard multi-core CPUs without the need for GPUs or specialized accelerators, highlighting its lightweight design. This efficiency ensures suitability for real-time or near real-time deployment in authentic educational settings, making it both practical and scalable for large-scale applications.

\section{Discussion}\label{sec:discussion}

The development of VisioPhysioENet introduces a novel approach in the area of multimodal engagement detection by leveraging both visual and physiological data streams. This innovative integration provides a more nuanced understanding of learner engagement, significantly improving both accuracy and robustness. With an accuracy of 63.09\% on the DAiSEE dataset, VisioPhysioENet surpasses existing unimodal and multimodal methods. A distinct advantage of VisioPhysioENet lies in its advanced feature extraction techniques and efficient data integration strategy, which effectively handles real-world educational environments. The use of Dlib and OpenCV for visual feature extraction and the POS method for physiological signal analysis also ensures high performance with minimal computational overhead, as further evidenced by the runtime analysis in Table~\ref{tab:comptime} and Fig.~\ref{fig:enter-label}, which shows substantial speed-ups even on standard CPUs. This combination of accuracy, efficiency, and unobtrusiveness reinforces the lightweight design goal, making the system feasible for deployment on standard, non-specialised hardware in real-life settings.

As shown in Tables \ref{tab:sota}, \ref{tab:mlvisual} and \ref{tab:mlphysio}, the visual modality achieves 43.90\% accuracy, while physiological signals alone yield 55.25\%. By integrating these two streams, VisioPhysioENet attains 63.09\%, illustrating how the multimodal approach captures complementary information often overlooked in unimodal systems. This combination leads to more robust performance and offers deeper insights into learners’ engagement states. Such synergy is particularly valuable in diverse educational settings where variations in gaze, facial expressions, and physiological responses contribute to the overall engagement profile.

An important design decision is the use of rPPG instead of contact PPG sensors. While contact PPG (cPPG) typically provides higher raw signal-to-noise ratio and supports richer heart rate variability metrics, it introduces deployment friction (wearable devices, hygiene, learner compliance) and limits scalability in classrooms \cite{kumar2023interpretable}. rPPG, especially the POS method adopted here, enables simultaneous, unobtrusive acquisition from standard webcams, avoiding behaviour alteration due to instrumentation. Prior comparative studies report low mean absolute heart-rate errors (often $\leq 3$--$5$ BPM under moderate motion) between POS-based rPPG and cPPG \cite{7565547,ghosh2015remote}, indicating that the reduced fidelity is still sufficient for coarse autonomic indicators correlated with attention and cognitive effort. Our quality gating (window rejection by motion/energy criteria) further narrows the performance gap, ensuring that retained rPPG segments yield heart-rate and peak-to-peak statistics that are discriminative enough to elevate classification when fused with visual cues.

Despite these strengths, the system faces challenges, especially regarding sensitivity to environmental variables such as lighting and motion artifacts. These factors can complicate data collection and introduce noise. Consequently, further refinement and additional testing are needed to ensure greater robustness across diverse populations and scenarios. Moreover, integrating physiological signals adds complexities in data capture and interpretation. The small changes in emotional or physical state can alter physiological readings. Improved algorithms are crucial to distinguish between the variations that genuinely indicate engagement and those arising from unrelated factors.

A second design choice concerns estimating head pose and gaze purely via computer vision rather than inertial measurement units (IMUs). IMUs can deliver high-rate, low-latency orientation with resilience to some visual occlusions, but they impose device cost, per-user calibration, potential discomfort, and classroom management overhead. Our landmark-driven solvePnP approach provides angular precision adequate for categorical engagement inference because decision thresholds (neutral bands, turn/tilt limits) are deliberately wider than typical landmark-induced jitter (a few degrees at common webcam resolutions) \cite{golob2022analysisfacedetectionface}. By mapping continuous pose and gaze to robust bins, we attenuate the effect of minor estimation noise while preserving meaningful attentional shifts (sustained yaw, downward pitch, lateral gaze).

Importantly, the pipeline is parallelised with a ThreadPoolExecutor for per-frame pre-processing, which, as shown in Table~\ref{tab:comptime}, reduces the processing time for a 64-frame batch from 732.6 seconds (sequential) to 293.0 seconds (parallel). This represents an approximate $2.5\times$ speed-up, sustaining around 22 fps which is well above the 5 fps acquisition rate, while leaving ample headroom for downstream inference. Such efficiency, achieved without GPUs or specialised accelerators, confirms that VisioPhysioENet is lightweight, scalable, and ready for real-world, large-scale deployment.

VisioPhysioENet’s strong performance in utilizing multimodal (visual and physiological) data suggests its applicability in various applications. For example, in education, it can monitor student engagement levels, enhancing instructional strategies and ultimately improving learning outcomes \cite{sharma2022student}. In corporate training, this approach could strengthen employee learning retention \cite{ma2022work}. Healthcare is another promising domain: by gauging patients’ understanding of treatment plans, clinicians can adjust patient education \cite{kumar2024measuring}. In advertisement, marketing, and consumer research, insights into consumer responses to media and advertising can be gleaned using this system \cite{john2020elaboration}. Additionally, virtual events and conferences can benefit, as real-time feedback can inform dynamic content adjustments to heighten attendee engagement \cite{vedernikov2024analyzing}. The characteristics of accuracy, efficiency, low hardware demands, and unobtrusive sensing position VisioPhysioENet as a practical solution for sustained, real-world use across domains.

\section{Conclusion} \label{sec:conc}
VisioPhysioENet, a novel multimodal system that integrates visual cues and physiological signals, has been proposed to effectively detect learner engagement. Achieving a commendable accuracy of 63.09\% on the DAiSEE dataset, it outperforms several state-of-the-art methods. This balance of accuracy and efficiency combined with its lightweight design and real-time processing capabilities makes it a robust solution for real-world applications, especially in dynamic learning environments. The system leverages Dlib and OpenCV for visual feature estimation and the POS method for physiological signals, enabling low-latency inference without compromising model performance. Its usability is further enhanced by an optimized preprocessing pipeline and ThreadPoolExecutor–based parallelism, ensuring smooth operation in live classroom monitoring or online education platforms without overwhelming compute resources.  
 
Despite these strengths, VisioPhysioENet remains sensitive to environmental variations such as lighting changes, camera angles, and sensor placement, and requires extensive validation in naturalistic settings to confirm its robustness and generalizability. Future enhancements will focus on improving scalability and robustness across diverse user groups and contexts. We will integrate additional modalities such as audio prosody, keystroke dynamics, and contextual metadata explore domain-adaptation strategies, and leverage distributed inference and edge-computing frameworks for real-time deployment at scale. Efforts will also target seamless incorporation into varied educational scenarios (online platforms, virtual classrooms, hybrid setups), ensuring adaptability, usability, and impact in personalized learning experiences. Through these developments, VisioPhysioENet aims to become a versatile, user-friendly tool that advances intelligent educational technologies.  


\section*{Acknowledgments}
The authors thankfully acknowledge IIT Hyderabad for releasing the DAiSEE dataset and the OS Lab at Computer Science and Engineering Department, NIT Jalandhar for providing the computational resources. This work was supported partially by the Eudaimonia Institute of the University of Oulu, Finland. We also extend our gratitude to collaborators from CMVS, University of Oulu and Zhejiang University for constructive discussions and reviewers whose feedback helped improve this work.

\section*{Declarations}
\noindent \textbf{Funding}: Not applicable.\vspace{.1in}

\noindent \textbf{Conflicts of interest}: Authors have no conflict of interest.\vspace{.1in}

\noindent \textbf{Code availability}: available at 
\href{github.com/MIntelligence-Group/VisioPhysioENet.}.\vspace{.1in}

\noindent \textbf{Availability of data and material}: Existing datasets have been used. Their details have been provided in this manuscript and above GitHub reposittory.\vspace{.1in}

\noindent \textbf{Authors' contributions}: 
\textit{Kanav Goyal}: Methodology, Implementation, Visual Representation.
\textit{Nischay Verma}: Data Curation, Experiments, Ablation Studies. 
\textit{Alakhsimar Singh}: Experiments, Validation, Result Analysis.
\textit{Puneet Kumar}: Writing - editing \& review, Mentoring and Supervision.
\textit{Xiaobai Li}: Conceptualization, Writing - review, Supervision.
\textit{Amritpal Singh}: Mathematical Representation, Writing - original draft \& editing, Project administration.\vspace{.1in}

\noindent \textbf{Ethics approval}: Not applicable.\vspace{.1in}

\noindent \textbf{Consent to participate}: Not applicable.\vspace{.1in}

\noindent \textbf{Consent for publication}: Not applicable.\vspace{.1in}

\noindent This article does not contain any studies with human participants or animals performed by any of the authors	

\bibliography{ref} 
\end{document}